%% file: Paper4132.tex
\def\year{2022}\relax
\documentclass[letterpaper]{article} 
\usepackage{aaai22}  
\usepackage{times}  
\usepackage{helvet}  
\usepackage{courier}  
\usepackage[hyphens]{url}  
\usepackage{graphicx} 
\usepackage{xcolor}
\usepackage{natbib}  
\usepackage{caption} 
\DeclareCaptionStyle{ruled}{labelfont=normalfont,labelsep=colon,strut=off} 
\usepackage{algorithm}
\usepackage{algorithmic}
\usepackage{multirow}
\usepackage{diagbox}
\usepackage{amssymb}
\usepackage{authblk}
\usepackage[colorlinks,bookmarks=false,citecolor=black]{hyperref}
\hypersetup{
    linkcolor=black,
    filecolor=black,      
    urlcolor=black,
    citecolor=black,
}

\usepackage{newfloat}
\usepackage{listings}
\floatstyle{ruled}
\newfloat{listing}{tb}{lst}{}
\floatname{listing}{Listing}

\def\ie{\textit{i.e.}}

\usepackage{amsmath}
\usepackage{multirow}      
\usepackage{multicol}

\title{Enhancing Pseudo Label Quality for Semi-Supervised  Domain-Generalized Medical Image Segmentation}

\author {
	\textbf{Huifeng Yao\textsuperscript{\rm 1},}
	\textbf{Xiaowei Hu\textsuperscript{\rm 2},}
    \textbf{Xiaomeng Li\textsuperscript{\rm 1,3}\thanks{Corresponding Author: {\tt\small eexmli@ust.hk}}}
}
\affiliations {
	\textsuperscript{\rm 1} Department of Electronic and Computer Engineering, The Hong Kong University of Science and Technology\\
	\textsuperscript{\rm 2} Department of Computer Science and Engineering, The Chinese University of Hong Kong\\
	\textsuperscript{\rm 3} The Hong Kong University of Science and Technology Shenzhen Research Institute\\
	{\tt\small hyaoad@connect.ust.hk, xwhu@cse.cuhk.edu.hk, eexmli@ust.hk}
}

\usepackage{bibentry}

\newif\ifshowcomments
\showcommentstrue

\ifshowcomments
\newcommand{\TODO}[1]{{\color{red}{[TODO: #1]}}}
\newcommand{\revised}[1]{{\color[rgb]{0.0,0.0,0.0}{#1}}}

\else
\newcommand{\TODO}[1]{}
\newcommand{\revised}[1]{}
\newcommand{\revised}[1]{}
\newcommand{\phil}[1]{}
\fi

\begin{document}

\maketitle

\input{abstract}

\input{introduction}

\input{related}
\input{method}

\input{experiment}

\input{conclusion}


\bibliography{ref}

\end{document}

%% file: abstract.tex
\begin{abstract}
Generalizing the medical image segmentation algorithms to unseen domains is an important research topic for computer-aided diagnosis and surgery. 
Most existing methods require a fully labeled dataset in each source domain. 
Although some researchers developed a semi-supervised domain generalized method, it still requires the domain labels.  
This paper presents a novel confidence-aware cross pseudo supervision algorithm for semi-supervised domain generalized medical image segmentation. The main goal is to enhance the pseudo label quality for unlabeled images from unknown distributions. 
To achieve it, we perform the Fourier transformation to learn low-level statistic information across domains and augment the images to incorporate cross-domain information. 
With these augmentations as perturbations, we feed the input to a confidence-aware cross pseudo supervision network to measure the variance of pseudo labels and regularize the network to learn with more confident pseudo labels. 
Our method sets new records on public datasets, \ie, M\&Ms and SCGM.
Notably, without using domain labels, our method surpasses the prior art that even uses domain labels by 11.67\% on Dice on M\&Ms dataset with 2\% labeled data. 
Code is available at
\href{https://github.com/XMed-Lab/EPL_SemiDG}{https://github.com/XMed-Lab/EPL\_SemiDG}.



\end{abstract}

%% file: introduction.tex
\section{Introduction}

\begin{figure}[]
\centering
\includegraphics[width=0.46\textwidth]{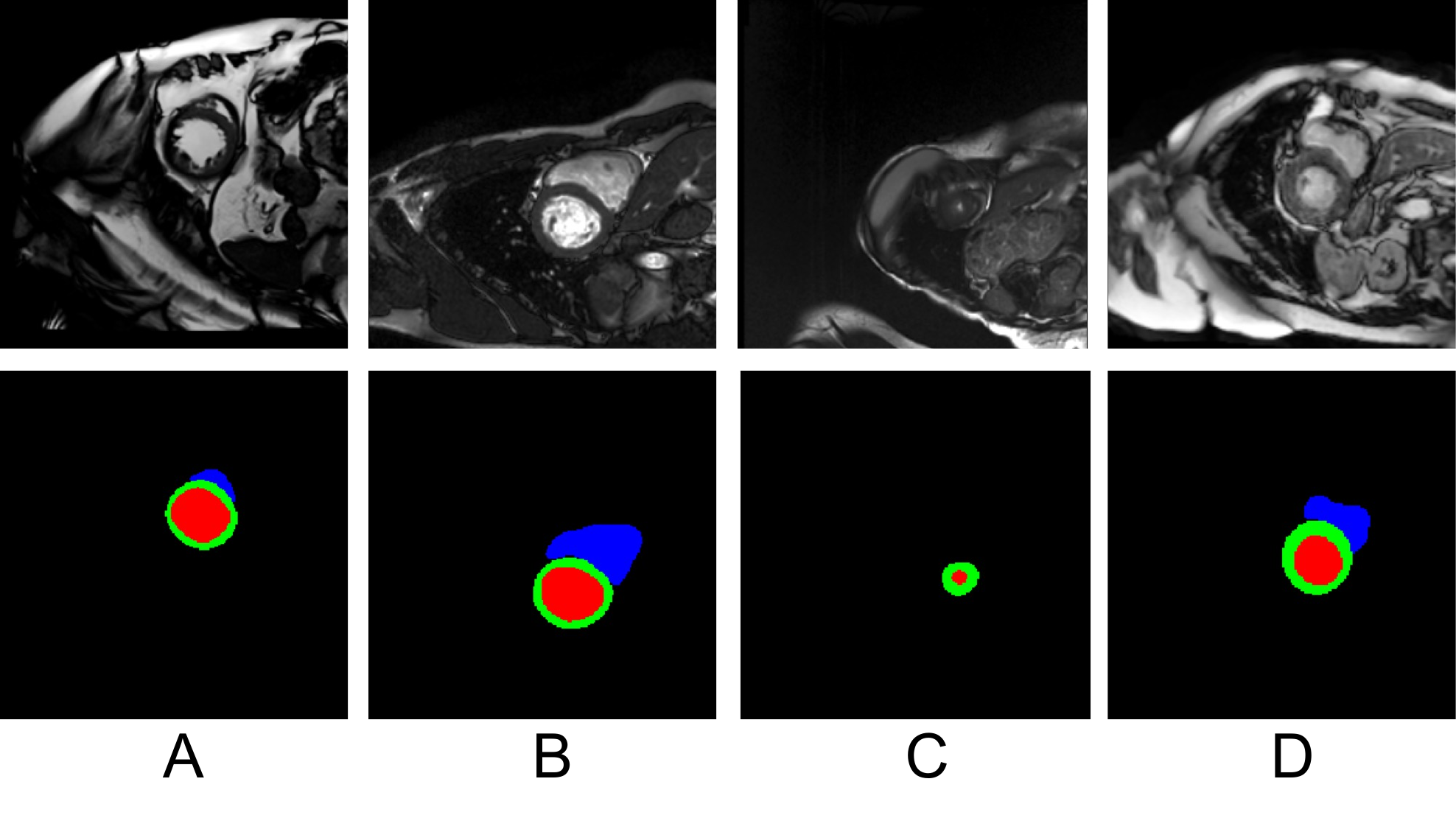} 
\caption{Some 2D slices of MRI from M\&Ms dataset. The first row is images from four different domains (A,B,C,D) and the second row is the corresponding segmentation masks. The red, blue, and green colors  refer to the left ventricle blood pool,the right ventricle blood pool, and the left ventricular myocardium, respectively.} 
\label{Fig.all_domain} 
\end{figure}

Medical image segmentation is one of the fundamental tasks in computer-aided diagnosis and computer-aided surgery. 
In recently years, researchers have developed many convolutional neural networks for medical image segmentation, such as U-Net~\cite{ronneberger2015u}, DenseUNet~\cite{li2018h}, nnUNet~\cite{isensee2021nnu}, and HyperDenseNet~\cite{dolz2018hyperdense}.
Medical images are usually collected from different clinical centers with different scanners~\cite{li2020self,puyol2021fairness,wang2020meta}.  
As a result, they may have apparent domain shifts due to variation in patient populations, scanners, and scan acquisition settings; see examples in Figure~\ref{Fig.all_domain}. 
However, the above methods generate inferior results when testing on images from unseen domains; see results in Table~\ref{tab1} and Table~\ref{tab2} in the experiments. Hence, it is crucial to strengthen the model’s generalization ability over different domain shifts. 

One naive solution to address the domain shifts is to obtain and annotate as much data as possible. However, the annotation cost is quite expensive to the community. 
Another solution is to train the model on the source domains and generalize it to the target domain with some information from the target domain, namely domain adaptation (DA)~\cite{bian2020uncertainty,pomponio2020harmonization}.
For example, \citet{pomponio2020harmonization} developed cross-site MRI harmonization
to enforce the source and target domains to share similar image-specific characteristics. 
Domain generalization (DG) is a more strict setting, where the difference with DA is that the model \emph{does not use any information from the target domain}. 
In this paper, we focus on this challenging but more practical problem, \ie, train the model on the source domains (A, B, and C) and test it on an unseen domain (D). 

Existing domain generalization methods have been developed for various medical image applications, such as cardiac segmentation~\cite{liu2020disentangled}, skin lesion classification~\cite{li2020domain}, spinal cord gray matter segmentation~\cite{li2020domain} and prostate MRI segmentation~\cite{liu2020shape}. 
For example, \citet{liu2020disentangled} proposed two data augmentation methods: ``Resolution Augmentaion'' and ``Factor-based Augmentation'' to generate more diverse data. 
\citet{li2020domain} proposed to learn a representative feature space through variational encoding to capture the semantic-consistent information among multiple domains. 
However, the above methods require that each domain should have fully-labeled datasets, and are not applicable if only partial data in each domain is labeled. 
Recently, \citet{liu2021semi} considered addressing this problem and presented a meta-learning approach with disentanglement for semi-supervised domain generalization. 
However, their method requires the source domain labels, which may not easily be obtained in clinical practice. 
Here, we consider a more practical setting for semi-supervised domain-generalized medical image segmentation: \emph{training data consists of labeled and unlabeled images from three source domains without knowing domain labels, and test data is from an unknown distribution}.


One straightforward solution is to directly use semi-supervised semantic segmentation methods~\cite{chen2021semi,Lai_2021_CVPR,Lee_2021_CVPR}. 
For example, \citet{chen2021semi} introduced a cross pseudo supervision method, where two segmentation networks are randomly initialized and supervised separately by the corresponding pseudo labels.
The unlabeled images in semi-supervised semantic segmentation usually are from the same distribution as the labeled one; thus, the pseudo labels can be directly used to refine another segmentation module. However, in our problem, the unlabeled images are from an unknown distribution, leading to a biased pseudo label. 

To this end, we introduce a novel confidence-aware cross pseudo supervision algorithm. The key idea is that two segmentation networks that shared the same structure are initialized differently. Hence, we can encourage consistent outputs from two networks. For the unlabeled images, each segmentation network can generate a pseudo label, which can be used as an additional signal to supervise the other segmentation network. 
To improve the quality of pseudo labels for unlabeled images from unknown domains, we propose to use Fourier transformation for unlabeled images, which can not only help obtain the low-level statistic information across different domains, but also augment the image samples by incorporating the cross-domain information.
Then, we develop the confidence-aware regularization to measure the pseudo variances generated by the original image and the image augmented by Fourier transformation. It helps improve the quality of pseudo labels, thus facilitating the learning process of the whole framework.

This paper has the following contributions:

\begin{itemize}
  \item We present a simple yet effective method for semi-supervised domain-generalized medical image segmentation. Our key idea is to enhance the quality of pseudo labels for unlabeled images from unknown distributions.
  
  \item We introduce two innovative modules: Fourier-transform-based data augmentation to augment the samples with cross-domain information, and confidence-aware cross pseudo supervision to measure the variances of pseudo labels.

  \item Our method achieves the new state-of-the-art performance on M\&Ms and SCGM dataset. Notably, without using domain labels, our method surpasses the prior art that even uses domain labels by 11.67\% on Dice on M\&Ms dataset with only 2\% labeled data. 
  
\end{itemize}

\begin{figure*}[tp] 
	\centering 
	\includegraphics[width=1\textwidth]{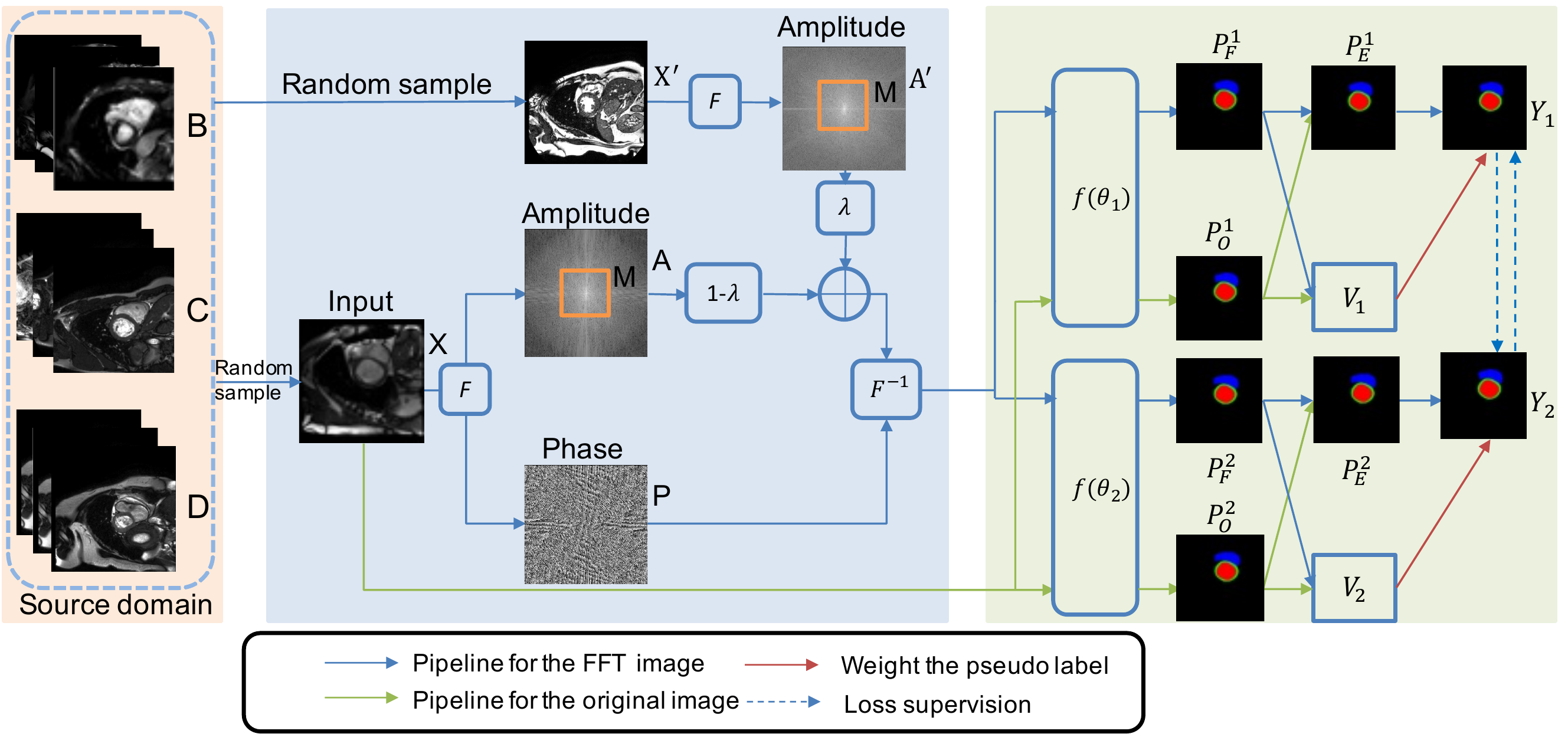}
	\caption{The overall architecture of the proposed confidence-aware cross-pseudo supervision network. B, C, and D are the source domains and A is the target domain. Note that the network does not use domain labels (\ie, B, C, and D) during training.  } 
	\label{Fig.pipeline} 
\end{figure*}

%% file: related.tex
\section{Related Work}
\subsection{Domain Generalization}
Domain generalization (DG) trains the model with multiple source domains and generalizes it to unseen target domains. Existing DG methods can be categorized into three classes: representation learning~\cite{zhou2020domain}, learning strategy design~\cite{yeo2021robustness}, and data manipulation~\cite{tobin2017domain}.
Representation learning methods mainly follow the idea of domain adaptation by learning domain-invariant features or explicitly feature alignment between domains. For example, \citet{zhou2020domain} aligned the marginal distribution of different source domains via optimal transport by minimizing the Wasserstein distance to achieve domain-invariant feature space. Learning strategy methods focus on exploiting the general learning strategy to promote the generalization capability. For example, \citet{yeo2021robustness} used the ensemble learning method to ensembling the middle domains into one strong prediction with uncertainty as to their weight. Data manipulation aims to manipulate the inputs to assist in learning general representations. For example, \citet{tobin2017domain} first used this method to generate more training data from the simulated environment for generalization in the real environment.

Recently, the Fourier transformation method has also proven to be very effective for DG. \citet{xu2021fourier} proposed a Fourier-transform-based framework for domain generalization by replacing the amplitude spectrum of a source image with that of another image from a different source domain. Their method achieves a remarkable performance in classification tasks. \citet{liu2021feddg} also used the
Fourier transformation method in federated learning and proves that it is a useful augmentation for medical image segmentation under federated learning. 

Unlike the above methods, we use the Fourier transformation as a data augmentation to get low-level statistic information among different domains and incorporate cross-domain information for unlabeled images. 
With these perturbations, we can measure the confidences of pseudo labels, and then enhance the model performance with reliable pseudo labels. 

\subsection{Semi-supervised Semantic Segmentation}
Unlike the image classification task, manually labeling pixel-wise annotations for the segmentation task is expensive and time-consuming. 
Existing methods for semi-supervised segmentation can be broadly classified into two categories: self-training~\cite{lee2013pseudo} and consistency learning~\cite{tarvainen2017mean,li2018semi, yu2019uncertainty, li2020transformation}.
The self-training method mainly uses the pseudo-label~\cite{lee2013pseudo} method to improve performance. This is initially developed for using unlabeled data in classification tasks. Then, it is also applied for semi-supervised segmentation~\cite{feng2020semi,ibrahim2020semi} recently. This method uses the pseudo segmentation maps of unlabeled data acquired by the previously trained segmentation model on labeled data to retrain the segmentation model. The process can be iterated multiple times.
Consistency learning encourages the model to have a similar output for the same input after different transformations. It imposes the consistency constraint between the predictions of different augmented images so that the decision function lies in the low-density region. Mean teacher~\cite{tarvainen2017mean} is a famous work of consistency learning in the classification task. Many works~\cite{kim2020structured,french2019semi} also use this in semi-supervised segmentation tasks. 
Recently, many methods combined these two ideas to get better performance in segmentation tasks. PseudoSeg~\cite{zou2020pseudoseg} performed both strong and weak augmentation on the same input images and used the weakly augmented image as the pseudo label. 
Cross pseudo supervision(CPS)~\cite{chen2021semi} also employed the pseudo-label method to expand the dataset and used consistency learning to learn a more compact feature encoding. 
Unlike these methods where unlabeled images are from the same domain, our method involves unlabeled images from an out-of-domain distribution, so the pseudo label generated by the network may not be reliable. Hence, we develop two innovative modules to enhance the quality of pseudo labels: one is Fourier-transform to augment cross-domain information for a single input from the source domain; the other is to measure the confidence of pseudo labels to get reliable supervision signals. 

%% file: method.tex
	\section{Methodology}
	
	%
	
	\subsection{Data Augmentation by Fourier Transformation}
	Figure~\ref{Fig.pipeline} illustrates the overall architecture of our method. 
	During the training process, we randomly take an image $X$ from source domain as the input image without knowing the domain label, and then perform the Fourier transformation $F$ to transfer the image to the frequency domain and obtain an amplitude spectrum $A$ and a phase image $P$, where the amplitude spectrum contains low-level statistics while the phase image includes high-level semantics of the original signal.
	To improve the capability of domain generalization, we randomly select another sample $X'$ from the source domain, perform the Fourier  transformation, and obtain another amplitude $A'$.
	Then, we augment the first image $X$ by incorporating the amplitude information of the second image $X'$ by:
	\begin{equation}
	\mathcal{A}_{new} \ = \ (1-\lambda) A *(1-\mathcal{M})+\lambda \mathcal{A}' * \mathcal{M} \ ,
	\end{equation}
	where $\mathcal{A}_{new}$ is the newly generated phase image; $\lambda$ is a parameter that used to adjust the ratio between the phase information of $X$ and $X'$; and $M$ is a binary mask to control the spatial range of amplitude spectrum to be exchanged and we set $M$ as the central region of the amplitude spectrum that contains low-frequency information.
	After that, we transform the merge sample from the frequency domain to the image domain through $F^{-1}$ to obtain the image sample $Z$ that augmented through Fourier transformation and incorporated the low-level information from another sample:
	\begin{equation}
	Z \ = \ \mathcal{F}^{-1}\left(\mathcal{A}_{new}, \mathcal{P}\right) .
	\end{equation}
	
	\subsection{Confidence-Aware Cross Pseudo Supervision}
	Next, we take both the original image $X$ and the transformed image $Z$ into two parallel segmentation networks $f(\theta_1)$ and $f(\theta_2)$, where these two networks have the same structure but their weights are initialized differently. 
	%
	
	\begin{figure}[!htb] 
    \centering 
    \includegraphics[width=0.48\textwidth]{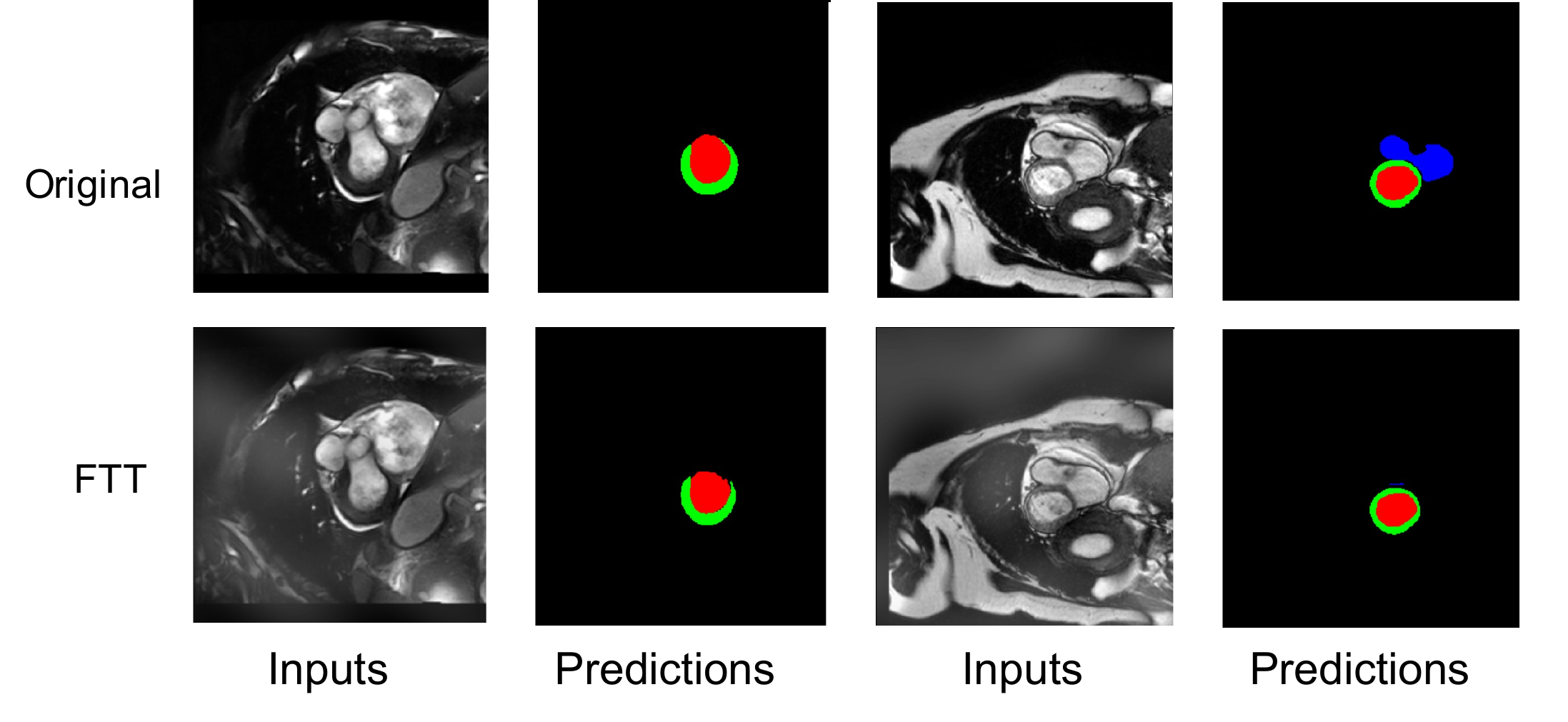} 
    \caption{These images show that the predictions of the original image and transformed image may have a large variance. Hence, we present the confidence-aware cross pseudo supervision mechanism to reduce the weight of training loss in such cases.}
    \label{Fig.variance} 
    \end{figure}
	
	For each segmentation network, we obtain the predictions of the the original image $X$ and the transformed image $Z$ by:
	\begin{equation}
	\begin{aligned}
	P_F^1 & = f(\theta_1)(Z), \\ 
	P_O^1 & = f(\theta_1)(X), \\ 
	P_F^2 & = f(\theta_2)(Z), \\ 
	P_O^2 & = f(\theta_2)(X).
	\end{aligned}
	\end{equation}
	%
	
	For the unsupervised part, since we do not have the supervision signals for the unlabeled data, after obtaining the predictions from different networks, we aim to adopt the prediction from one network to supervise the prediction from another network. %
	This technique is proposed by~\cite{chen2021semi} and named as the cross pseudo supervision.
	However, in~\cite{chen2021semi}, the label and unlabeled data are from the same domain with less variation, hence, the generated pseudo labels usually have a high quality.
	In contrast, the data in our task is from multiple domains and the large variance in the training samples from different domains may lead to low-quality pseudo labels.
	To improve the label quality and reduce the influence of low-quality labels, in this work, we present a confident-aware cross pseudo supervision mechanism.

	%
	As shown in Figure~\ref{Fig.pipeline}, after obtaining the predictions $P_O^1$ and $P_F^1$ from the original image and transformed image, we compute the average value of $P_O^1$ and $P_F^1$ as the ensemble prediction result $P_E^1$:
	\begin{equation}
	P_E^1 \ = \ ( P_O^1 + P_F^1 ) / 2 ,
	\end{equation}  
	and we obtain the ensemble prediction result $P_E^2$ of the second network by:
	\begin{equation}
	P_E^2 \ = \ ( P_O^2 + P_F^2 ) / 2 .
	\end{equation}  
	To selectively use the pseudo labels as the supervisions and reduce the low-quality labels, we first compute the variance of the predictions of the original image and transformed image to measure the quality of the pseudo labels.
	The variance is computed as the KL-divergence of these two predictions:
	\begin{equation}
	\begin{aligned}
	V_1 & = E[ \ P_F^1 \ log(\frac{P_F^1}{P_O^1}) \ ] \ ,\\
	V_2 & = E[ \ P_F^2 \ log(\frac{P_F^2}{P_O^2}) \ ] \ ,
	\end{aligned}
	\end{equation}
	where $E$ compute the expectation value. 
	If the difference between these two predictions is large, the computed variance value is also large, which reflects these two prediction have a relative low quality, and vice versa. Figure~\ref{Fig.variance} shows some visual examples, where the predictions between the original image and the transformed image have a large variance.
	
	After obtaining the variances of the prediction of these two network, we formulate the confidence-aware loss function $L_{cacps} \ = \ L_a + L_b$ to optimize the networks by using the cross supervision signals:
	\begin{equation}
	\begin{aligned}
	L_a & = E[ \ e^{-V_1} L_{ce}(P_E^2, Y_1) + V_1 \ ], \\
	L_b & = E[ \ e^{-V_2} L_{ce}(P_E^1, Y_2) + V_2 \ ],
	\end{aligned}
	\end{equation}
	where $Y_1$ and $Y_2$ are the one-hot vectors generated from the probability maps $P_E^1$ and $P_E^2$, and $L_{ce}$ denotes the cross-entropy loss.
	
	For the supervised part, we use dice loss as loss function. We define the supervision loss as $L_s$.
	
	\begin{equation}
    L_s=E[L_{Dice}(P_O^1,G_1) + L_{Dice}(P_O^2,G_2)],
    \end{equation}
    where $L_{Dice}$ is the dice loss function and $G_1$($G_2$) is the ground truth.
    
    So, the whole training objective is written as:
    \begin{equation}
    \label{loss_final}
    L=L_{s} + \beta * L_{cacps},
    \end{equation}
    where $\beta$ is the CACPS weight. Its goal is to put the two losses into balance.
    
    During the test procedure, we use the ensemble of two models’ predictions as the final results.

%% file: experiment.tex
\section{Experimental Results}

\begin{table*}[tp]
\centering
\setlength{\tabcolsep}{11.5pt} 
\begin{tabular}{l|l|l|l|l|l}
\hline 
\diagbox{Method}{Target}  & \begin{tabular}{c}A\end{tabular}        & \begin{tabular}{c}B\end{tabular}         & \begin{tabular}{c}C\end{tabular}         & \begin{tabular}{c}D\end{tabular}         & Average   \\ \hline
nnUNet \cite{isensee2021nnu} & 65.30±17          & 79.73±10            & 78.06±11            & 81.25±8.3           & 76.09±6.3          \\ \hline
SDNet+Aug \cite{liu2020disentangled} & 71.21±13      & 77.31±10            & 81.40±8.0           & 79.95±7.8           & 77.47±3.9           \\ \hline
LDDG \cite{li2020domain} & 66.22±9.1            & 69.49±8.3          & 73.40±9.8           & 75.66±8.5           & 71.29±3.6           \\ \hline
SAML \cite{liu2020shape} & 67.11±10            & 76.35±7.9           & 77.43±8.3           & 78.64±5.8           & 74.88±4.6           \\ \hline
Meta \cite{liu2021semi} & 72.40±12            & 80.30±9.1           & 82.51±6.6           & 83.77±5.1           & 79.75±4.4           \\ \hline
Ours & \textbf{83.3±5.83} & \textbf{85.04±6.49} & \textbf{87.14±4.74} & \textbf{87.38±4.49} & \textbf{85.72±1.67} \\ \hline
\end{tabular}
\caption{Dice (\%) results and the standard deviations on M\&Ms dataset using 5\% labeled data. For “SDNet+Aug”, "Meta", and our method, the training data contains all unlabeled data and 5\% of labeled data from the source domains. The other models are trained by using 5\% labeled data only. Bold numbers denote the best performance.}
\label{tab1}
\end{table*}

\begin{table*}[!htb]
\centering
\setlength{\tabcolsep}{11.5pt} 
\begin{tabular}{l|l|l|l|l|l}
\hline
\diagbox{Method}{Target}  & \begin{tabular}{c}A\end{tabular}        & \begin{tabular}{c}B\end{tabular}         & \begin{tabular}{c}C\end{tabular}         & \begin{tabular}{c}D\end{tabular}         & Average   \\ \hline
nnUNet~\cite{isensee2021nnu} & 52.87±19            & 64.63±17           & 72.97±14           & 73.27±11           & 65.94±8.3           \\ \hline
SDNet+Aug~\cite{liu2020disentangled} & 54.48±18       & 67.81±14           & 76.46±12           & 74.35±11            & 68.28±8.6  \\ \hline
LDDG~\cite{li2020domain} & 59.47±12            & 56.16±14           & 68.21±11           & 68.56±10           & 63.16±5.4           \\ \hline
SAML~\cite{liu2020shape} & 56.31±13            & 56.32±15           & 75.70±8.7           & 69.94±9.8           & 64.57±8.5           \\ \hline
Meta~\cite{liu2021semi} & 66.01±12            & 72.72±10           & 77.54±10           & 75.14±8.4           & 72.85±4.3           \\ \hline
Ours & \textbf{82.35±6.24}  & \textbf{82.84±7.59} & \textbf{86.31±5.47} & \textbf{86.58±4.78} & \textbf{84.52±1.94} \\ \hline
\end{tabular}
\caption{Dice (\%) results and the standard deviations on M\&Ms dataset using 2\% labeled data. For “SDNet+Aug”, "Meta", and our method, the training data contains all the unlabeled data and 2\% of labeled data from source domains. The other models are trained by using 2\% labeled data only. Bold numbers denote the best performance.}
\label{tab2}
\end{table*}

\begin{table*}[!htb]
\centering
\setlength{\tabcolsep}{11.5pt} 
\begin{tabular}{l|l|l|l|l|l}
\hline
\diagbox{Method}{Target}  & \begin{tabular}{c}1\end{tabular}        & \begin{tabular}{c}2\end{tabular}         & \begin{tabular}{c}3\end{tabular}         & \begin{tabular}{c}4\end{tabular}         & Average   \\ \hline
nnUNet~\cite{isensee2021nnu} & 59.07±21            & 69.94±12           & 60.25±7.2           & 70.13±4.3           & 64.85±5.2           \\ \hline
SDNet+Aug~\cite{liu2020disentangled} & 83.07±16       & 80.01±5.2           & 58.57±10           & 85.27±2.2            & 76.73±11  \\ \hline
LDDG~\cite{li2020domain} & 77.71±9.1            & 44.08±12           & 48.04±5.5           & 83.42±2.7           & 63.31±17           \\ \hline
SAML~\cite{liu2020shape} & 78.71±25           & 75.58±12           & 54.36±7.6           & 85.36±2.8          & 73.50±12          \\ \hline
Meta~\cite{liu2021semi} & \textbf{87.45±6.3}            & 81.05±5.2           & 61.85±7.3           & 87.96±2.1          & 79.58±11           \\ \hline
Ours & 87.13±1.4  & \textbf{87.31±2.02} & \textbf{78.75±9.44} & \textbf{91.73±1.28} & \textbf{86.23±4.69} \\ \hline
\end{tabular}
\caption{Dice (\%) results and the standard deviations on SCGM dataset using 20\% labeled data. For “SDNet+Aug”, "Meta", and our method, the training data contains all the unlabeled data and 20\% of labeled data from source domains. The other models are trained by using 20\% labeled data only. Bold numbers denote the best performance.}
\label{tab3}
\end{table*}

\subsection{Implementation Details}
We implemented the model on Pytorch1.8 and trained it by using two NVidia 3090 GPUs with 377GB RAM on the Ubuntu20.04 system.
We implemented two segmentation networks by using DeepLabv3+~\cite{chen2017rethinking} with ResNet50~\cite{he2016deep} backbone, and adopted the weights trained on ImageNet~\cite{deng2009imagenet} for classification to initialize the parameters of the backbone network and other layers were initialized by random noise.
For M\&Ms dataset, we leveraged AdamW to optimize the network with the weight decay of 0.1, the learning rate of $0.0001$, and the batch size of 32. We trained the whole architecture for 20 epochs and the images were cropped to $288\times288$. We set $\beta$ in equation~\ref{loss_final} as three to balance the supervision loss and our proposed CACPS loss. We set $\lambda$ in equation 1 as 1.
For SCGM dataset, we leveraged AdamW to optimize the network with the weight decay of 0.1, the learning rate of $0.0001$, and the batch size of eight. We trained the whole architecture for 50 epochs and the images were cropped to $288\times288$. We set $\beta$ in equation~\ref{loss_final} as $1.5$ to balance the supervision loss and our proposed CACPS loss.  We set $\lambda$ in equation 1 as 0.8.
We also adopt the random rotation, random scaling, random crop, and random flip as the data augmentation strategies. 



\subsection{Datasets and Evaluation Metrics}
We adopt the multi-centre, multi-vendor \& multi-disease cardiac image segmentation (M\&Ms) dataset~\cite{campello2021multi} to evaluate of our method.
It contains 320 subjects, which are scanned at six clinical centres in three different countries by using four different magnetic resonance scanner vendors, \ie, Siemens, Philips, GE, and Canon, and we consider the subjects scanned from different vendors are from different domains (domains A,B,C,D).
For each subject, only the end-systole and end-diastole phases are annotated. 
The resolutions of the voxels range from 0.85×0.85×10 mm to 1.45×1.45×9.9 mm. 
In total, there are 95 subjects in domain A, 125 subjects in domain B, 50 subjects in domain C, and another 50 subjects in domain D.

We also adopt the spinal cord gray matter segmentation (SCGM) dataset\cite{prados2017spinal} to evaluate of our method. 
This dataset contains single channel Spinal Cord MRI data with gray matter labels from four different centers. Data is collected from four centers (UCL, Montreal, Zurich, Vanderbilt) using three different MRI systems (Philips Acheiva, Siemens Trio, Siemens Skyra) with institution specific acquisition parameters. So it has four domains (domains A,B,C,D) in total. The voxel resolutions range from 0.25 × 0.25 × 2.5 mm to 0.5 × 0.5 × 5 mm. Each domain has 10 labeled subjects and 10 unlabelled subjects.

\begin{figure}[tp] 
\centering 
\includegraphics[width=0.48\textwidth]{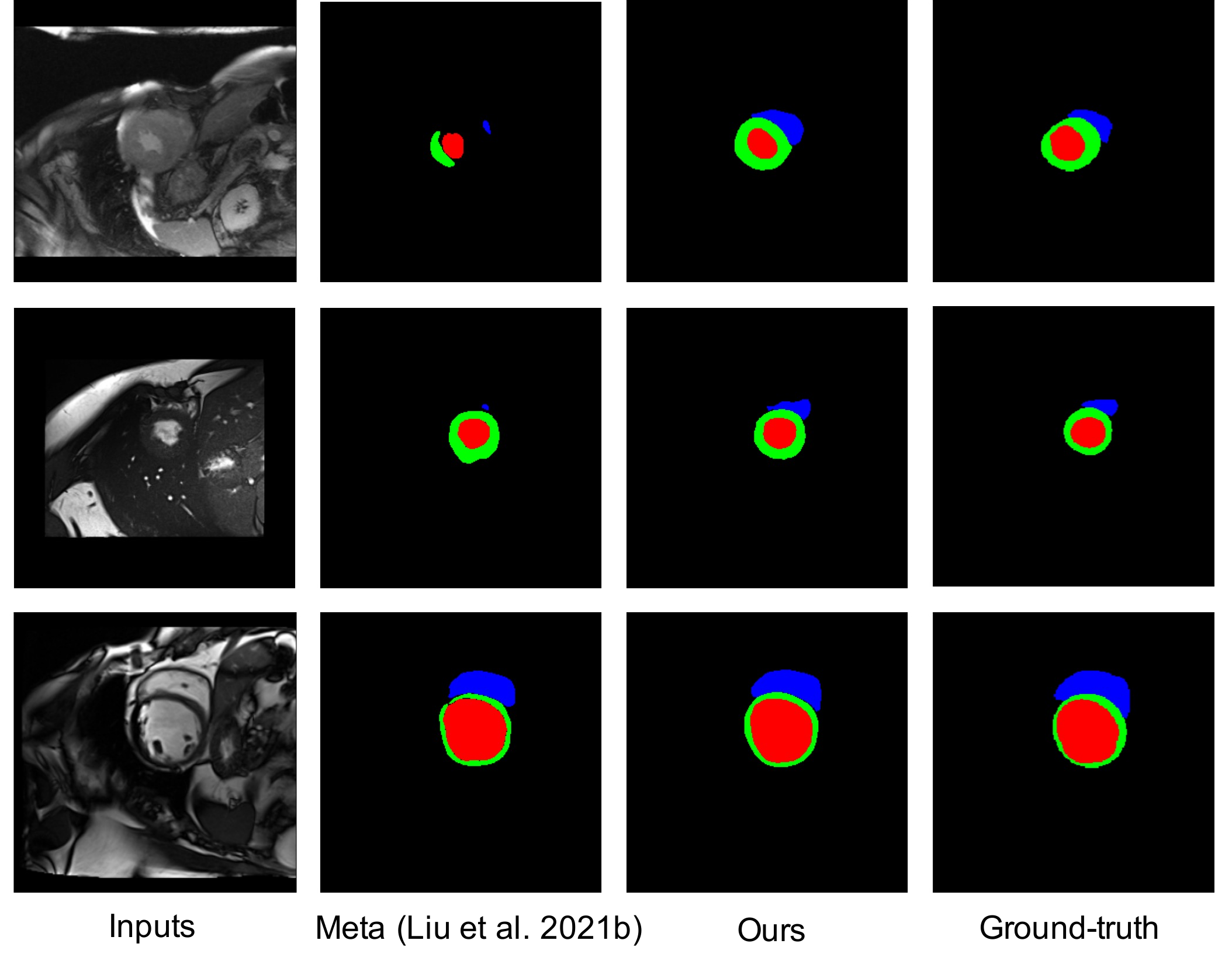} 
\caption{Visual comparison results on the M\&Ms dataset.} 
\label{Fig.compare} 
\end{figure}


We evaluate the segmentation performance by using Dice (\%) score, which is defined as:
\begin{equation}
Dice\left(P, G\right) = \frac{2 \times \left\vert P \cap G \right\vert}{\left\vert P \right\vert + \left\vert G \right\vert}  ,
\end{equation}
where $P$ and $G$ are the predicted segmentation result and ground truth  image, respectively. $\left\vert P \cap G \right\vert$ denotes the overlapped region between $P$ and $G$, while $\left\vert P \right\vert + \left\vert G \right\vert$ represents the union region.
In general, a better segmentation result has a larger $Dice$.


\subsection{Experiments on the M\&Ms Dataset}

\subsubsection{Comparison with the State-of-the-art Methods.}
We compare our method with the following state-of-the-art methods:

(i) \textbf{nnUNet~\cite{isensee2021nnu}}: is a self-adapting framework based on 2D and 3D U-Nets\cite{ronneberger2015u} for fast and effective segmentation. But it is not designed for domain generalization.

(ii) \textbf{SDNet+Aug~\cite{liu2020disentangled}}: first generates diverse data by rescaling the images to different resolutions within a range spanning different scanner protocols, and then generates more diverse data by projecting the original samples onto disentangled latent spaces and combining the learned anatomy and modality factors from different domains.

(iii) \textbf{LDDG~\cite{li2020domain}}: presents a method to learn a representative feature space through variational encoding with a novel linear-dependency regularization term to capture the shareable information among medical data, which is collected from different domains. It is the latest state-of-the-art model for domain-generalized medical image analysis in a fully supervised setting.

(iv) \textbf{SAML~\cite{liu2020shape}}: is a gradient-based meta-learning approach, which constrains the compactness and smoothness properties of segmentation masks across meta-train and meta-test sets in a fully supervised manner.

(v) \textbf{Meta~\cite{liu2021semi}}: is a semi-supervised meta-learning framework, which models the domain shifts by using the disentanglement and extracts robust anatomical features for predicting segmentation masks in a semi-supervised manner by applying multiple constraints with the gradient-based meta-learning approach. It is the latest state-of-the-art model for domain-generalized medical image analysis in a semi-supervised setting.

\subsubsection{Results on the M\&Ms dataset}

Table~\ref{tab1} and Table~\ref{tab2} report the comparison results on the M\&Ms dataset, where our method achieves the best performance on all different settings.
Specifically,  our method shows the improvements $\approx$ 6\% and $\approx$ 12\% in terms of Dice scores on the settings of 5\% labeled data and 2\% labeled data, respectively, by comparing with the previous best method.
%
Note that by comparing with other methods, our method gives a significant improvement on the setting that we adopt the image samples from domains B, C, D as the training data, and evaluate the trained model on the domain A. 
This is because the image samples in domain A are less similar to other domain, as also shown in Figure~\ref{Fig.all_domain}, and our method has a great generalization capability.

Figure~\ref{Fig.compare} shows more visual results by comparing with the previous best method, \ie, Meta~\cite{liu2021semi}.
From the results, we can see that our method generates the results that are more consistent with the ground truth images, while Meta~\cite{liu2021semi} tend to mis-segment some un-obvious regions of the input images.

\begin{figure}[tp] 
\centering 
\includegraphics[width=0.48\textwidth]{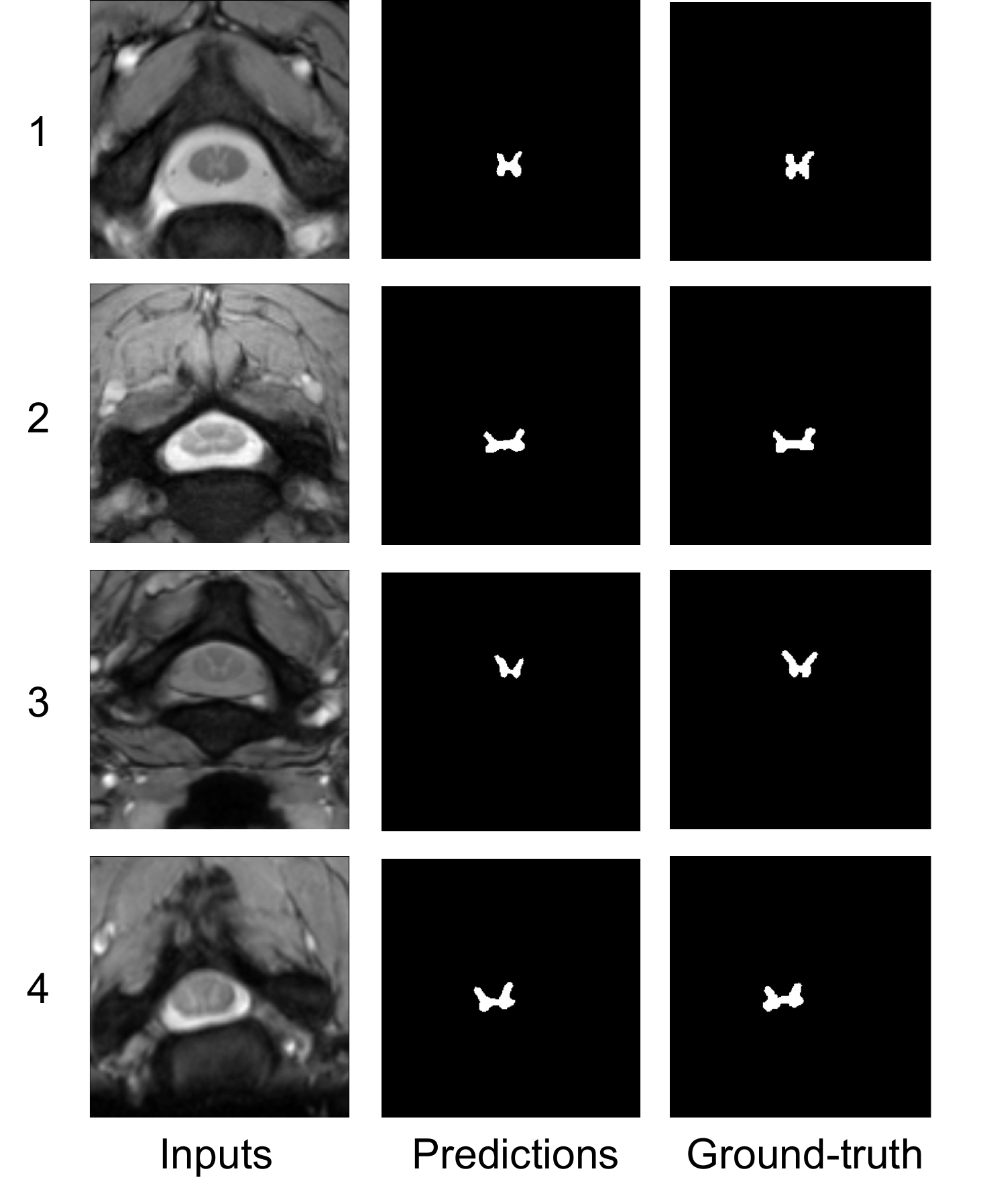} 
\caption{Visual comparison results on the SCGM dataset. We take the images from four different domains as the example visualization.} 
\label{Fig.scgm} 
\end{figure}

\subsubsection{Results on the SCGM dataset}
Table~\ref{tab3} reports the comparison results on the SCGM dataset, where our method shows the improvement $\approx$ 7\% in terms of Dice score on the setting of 20\% labeled data, by comparing with the previous best method.
Figure~\ref{Fig.scgm} shows the visual results produced by our method in different domains, where we can see that our method successfully generates the prediction results that are consistent with the ground truth images on all the four domains.

\begin{figure}[tp] 
\centering 
\includegraphics[width=0.5\textwidth]{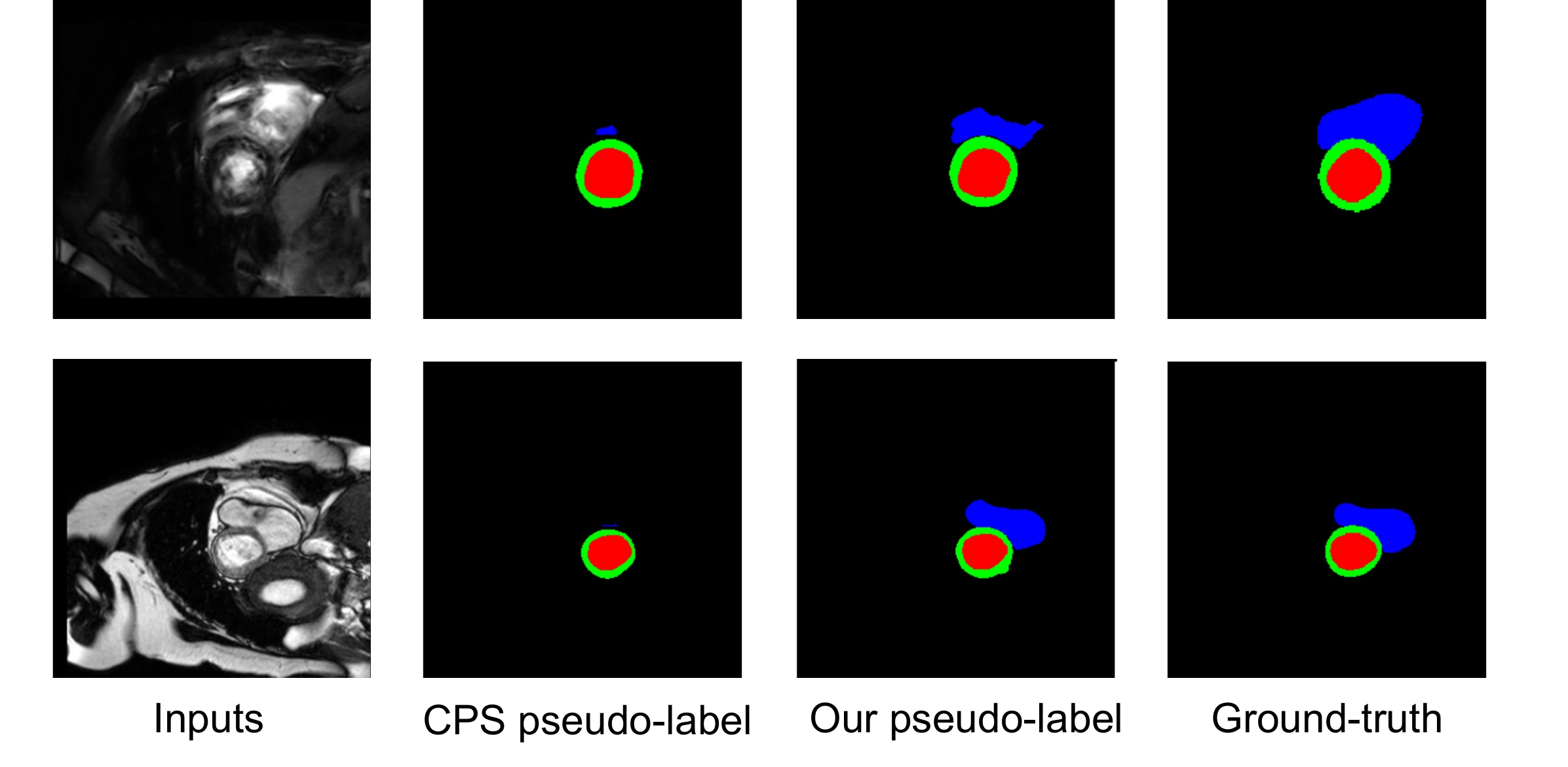} 
\caption{We compare the pseudo labels generated by CPS~\cite{chen2021semi} and by our method, where our method can generate the pseudo labels that are more consistent with the ground-truth images.} 
\label{Fig.pseudo_label} 
\end{figure}

\subsubsection{Ablation Study}
We perform an ablation study to evaluate the effectiveness of each component of our proposed method on the M\&Ms dataset.
First, we build our first baseline by using CPS~\cite{chen2021semi} designed for semi-supervised learning to perform segmentation task on different domains.
Then, we construct the second model by further using the Fourier transformation to perform data augmentation and further use CPS to train the network.
Third, we consider the confidence-aware cross pseudo supervision (CACPS) instead of simple CPS to optimize the network, which is the full pipeline of our proposed method.
Moreover, we visually compare the pseudo labels generated by CPS and our method in Figure ~\ref{Fig.pseudo_label}, where we can see that our pseudo labels are more consistent with the ground truth images.
Table~\ref{tab4} reports the results, where we can see that each component gives a clear improvement on the results.

\begin{table}
\centering
\caption{Ablation Study on the M\&Ms dataset by using $2\%$ labeled data. We consider domain A as the target domain and others as the source domain.}
\label{tab4}
\begin{tabular}{cccc} 
\hline
CPS & Fourier & CACPS & Dice(\%)    \\ 
\hline
\checkmark   &         &           & 80.46±4.67                               \\
\checkmark   & \checkmark      &           &81.47±4.66                                 \\
\checkmark   & \checkmark      & \checkmark        & 82.61±4.15                               \\
\hline
\end{tabular}
\end{table}

%% file: conclusion.tex
\section{Conclusion}

This paper presents a semi-supervised domain generalization method for medical images by formulating two novel techniques.
One is the Fourier-transform-based data augmentation and another is the confidence-aware cross pseudo supervision.
The Fourier-transform-based data augmentation helps to obtain the low-level statistic information from different domains, which augments the image samples by incorporating the cross-domain information.
The confidence-aware cross pseudo supervision algorithm measures the variance between the original image sample and the image sample augmented by Fourier transformation, which helps to improves the quality of pseudo labels, thus facilitating the learning process of the whole framework.
Finally, we evaluate our method on two public benchmark datasets, compare our method with various methods, and show its superiority over the other state-of-the-art methods.
In the future, we will jointly leverage the semi-supervised and weakly-supervised algorithms to improve the performance of medical image segmentation by exploring the knowledge of unlabeled data and the data with weaker labels, such as bounding boxes.
Moreover, we will explore our framework for more medical image segmentation tasks and integrate the network into the artificial intelligent systems on medical diagnostics.

\section{Acknowledgments} This work was supported by a research grant from Shenzhen Municipal Central Government Guides Local Science and Technology Development Special Funded Projects (2021Szvup139) and a research grant from HKUST Bridge Gap Fund (BGF.027.2021).